\documentclass{article}

\usepackage{microtype}
\usepackage{graphicx}
\usepackage{subcaption}
\usepackage{booktabs} % for professional tables

\usepackage{hyperref}

\usepackage[preprint]{icml2026}

\usepackage{amsmath}
\usepackage{amssymb}
\usepackage{mathtools}
\usepackage{amsthm}

\usepackage[capitalize,noabbrev]{cleveref}

\theoremstyle{plain}

\theoremstyle{definition}

\theoremstyle{remark}

\usepackage[textsize=tiny]{todonotes}

\begin{document}

\twocolumn[
  \icmltitle{\textsc{SciPaths}: Forecasting Pathways to Scientific Discovery}

  % It is OKAY to include author information, even for blind submissions: the
  % style file will automatically remove it for you unless you've provided
  % the [accepted] option to the icml2026 package.

  % List of affiliations: The first argument should be a (short) identifier you
  % will use later to specify author affiliations Academic affiliations
  % should list Department, University, City, Region, Country Industry
  % affiliations should list Company, City, Region, Country

  % You can specify symbols, otherwise they are numbered in order. Ideally, you
  % should not use this facility. Affiliations will be numbered in order of
  % appearance and this is the preferred way.

\icmlsetsymbol{equal}{*}

\begin{icmlauthorlist}
  \icmlauthor{Eric Chamoun}{cam}
  \icmlauthor{Yizhou Chi}{cam}
  \icmlauthor{Yulong Chen}{equal,cam}
  \icmlauthor{Rui Cao}{equal,sutd}
  \icmlauthor{Zifeng Ding}{equal,cam}
  \icmlauthor{Michalis Korakakis}{turing}
  \icmlauthor{Andreas Vlachos}{cam}
\end{icmlauthorlist}

\icmlaffiliation{cam}{University of Cambridge, Cambridge, United Kingdom}
\icmlaffiliation{turing}{The Alan Turing Institute, London, United Kingdom}
\icmlaffiliation{sutd}{Singapore University of Technology and Design, Singapore}

\icmlcorrespondingauthor{Eric Chamoun}{ec806@cam.ac.uk}

  % You may provide any keywords that you find helpful for describing your
  % paper; these are used to populate the "keywords" metadata in the PDF but
  % will not be shown in the document
  \icmlkeywords{Machine Learning, ICML}

  \vskip 0.3in
]
\printAffiliationsAndNotice{\icmlEqualContribution}

% this must go after the closing bracket ] following \twocolumn[ ...

% This command actually creates the footnote in the first column listing the
% affiliations and the copyright notice. The command takes one argument, which
% is text to display at the start of the footnote. The \icmlEqualContribution
% command is standard text for equal contribution. Remove it (just {}) if you
% do not need this facility.

% Use ONE of the following lines. DO NOT remove the command.
% If you have no special notice, KEEP empty braces:

% Or, if applicable, use the standard equal contribution text:
% \printAffiliationsAndNotice{\icmlEqualContribution}

\begin{abstract}

Scientific progress depends on sequences of enabling contributions, yet existing AI4Science benchmarks largely focus on citation prediction, literature retrieval, or idea generation rather than the dependencies that make progress possible. In this paper, we introduce \emph{discovery pathway forecasting}: given a target scientific contribution and the prior literature available at a specified time, the task is to (1) identify the enabling contributions required to realize it and (2) ground each in prior work when such prior work exists. We present \textsc{SciPaths}, a benchmark of 262 expert-annotated gold pathways and 2,444 silver pathways constructed from machine learning and natural language processing papers, where each pathway records enabling contributions, roles, rationales, and prior-work groundings or unmapped decisions. Evaluating frontier and open-weight language models, we find that the best model reaches only 0.189 F1 under strict semantic matching, with core methodological dependencies hardest to recover. Prior-work grounding improves substantially when gold enabling contributions are provided, showing that decomposition quality is a major bottleneck for end-to-end pathway recovery. \textsc{SciPaths} therefore shifts evaluation toward a missing capability in scientific forecasting: reasoning backward from a target contribution to the enabling scientific building blocks and prior-work dependencies that make it feasible.

\end{abstract}

\section{Introduction}
Scientific discoveries rarely arise in isolation: they build on enabling contributions and, in turn, enable subsequent work \citep{fortunato2018,uzzi2013,wu2019}. This raises a central question for scientific forecasting: \textit{given a target contribution, which enabling contributions are required to realize it?}

Two lines of work explore this question indirectly. Metascience studies of method recombination, concept prerequisites, and knowledge precedence describe how knowledge evolves \citep{chen,zhu2022,xiang2026}, but typically operate retrospectively over papers, concepts, or aggregate patterns. AI4Science systems support literature analysis, hypothesis generation, and idea evaluation \citep{reddy2024,boiko2023,cimon}, but usually treat ideas as standalone outputs rather than reasoning over the dependencies that make them feasible.
\begin{figure*}[t]
    \centering
    \includegraphics[width=\linewidth]{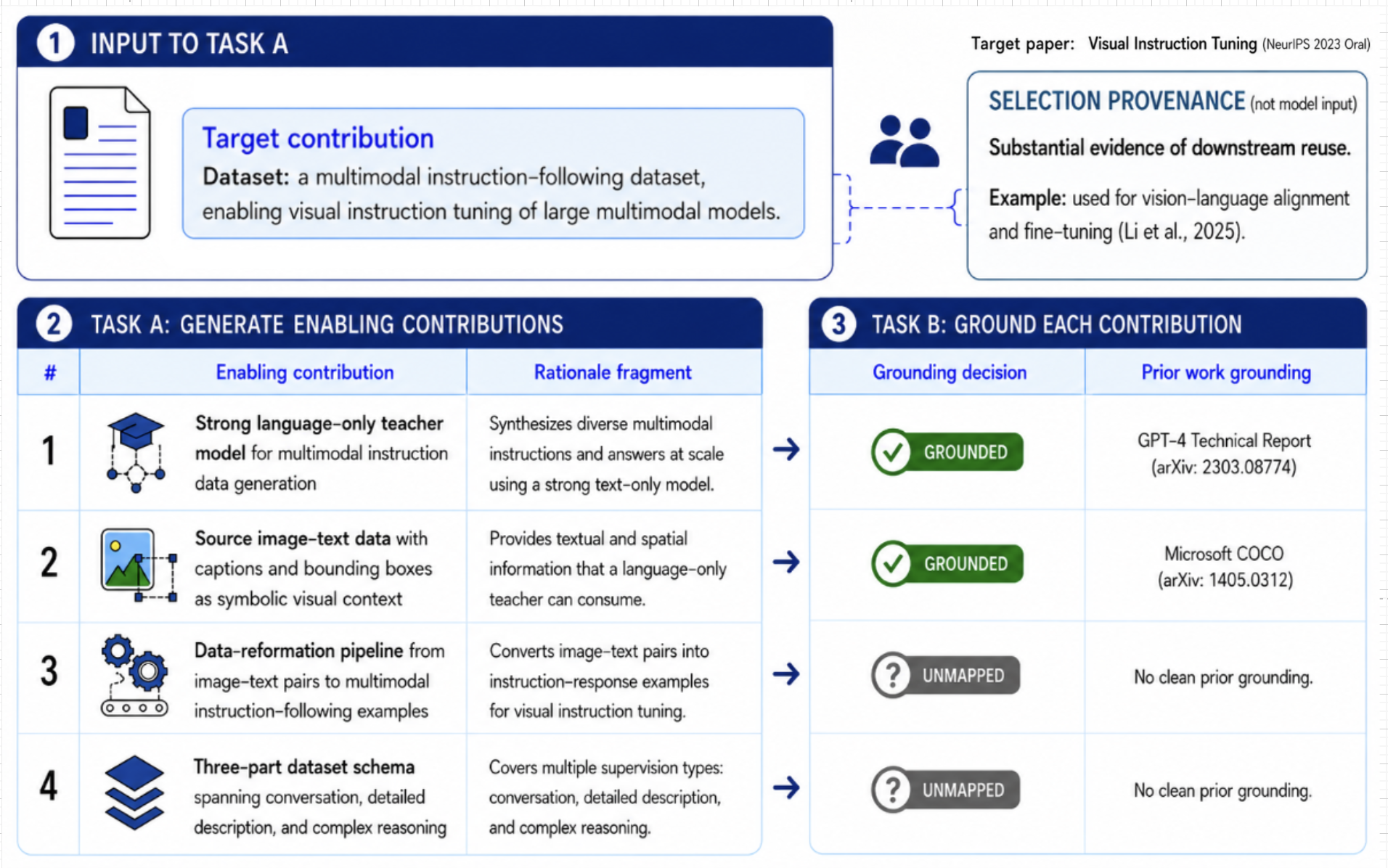}
\caption{
Example \textsc{SciPaths} instance and task structure. In the main Task A setting, the model receives a target contribution claim and predicts the enabling contributions required to realize it, along with rationale fragments. Selection provenance explains why the target contribution was included but is not provided as model input. Task B grounds each enabling contribution in prior work or marks it as unmapped. Rationale fragments are abbreviated for readability.
}
    \label{fig:example}
\end{figure*}
Closest to our work are citation-based formulations of scientific forecasting and citation recommendation. For example, \textsc{PreScience} \citep{prescience} predicts the key references that a target paper's authors are likely to build upon when creating a new contribution. However, citation-based supervision relies on influence proxies and operates at the level of whole papers, which can conflate heterogeneous contributions. Using the example from Figure~\ref{fig:example}, \textit{Visual Instruction Tuning} \citep{visual-it} releases both an instruction-tuning dataset and a general-purpose multimodal assistant; prior work essential for one contribution may be irrelevant to the other. A flat paper-level reference set cannot express which reference supports which target contribution or what enabling role it plays. Exact citation matching can also penalize models that identify the right enabling contribution but cite a different valid paper that could play the same role.

Complementary to citation-based forecasting, GIANTS \citep{giants} generates downstream insights from known parent papers, assuming the relevant prior work is already given. We target the missing dependency-identification step at finer granularity through three design decisions: (1) representing targets at the contribution level rather than the paper level, (2) separating enabling contributions from their prior-work grounding, and (3) evaluating models based on contribution-level correctness rather than exact citation matching.

We introduce \textsc{SciPaths}, a benchmark for \textit{discovery pathway forecasting}. Given a target contribution, the task is to (a) identify the enabling contributions required to realize it and (b) ground each one in representative prior work when such prior work exists, or mark it as unmapped (Figure~\ref{fig:example}). This separates \textit{what is needed} from \textit{which prior work realizes it}. We construct \textsc{SciPaths} from machine learning and natural language processing papers by selecting target contributions with evidence of downstream reuse; for example, the instruction-tuning dataset in Figure~\ref{fig:example} was later used for vision-language alignment and fine-tuning. This criterion focuses the benchmark on contributions that became actionable building blocks for subsequent research, rather than on all claims made in a paper.

Expert annotators validate each target contribution and decompose it into a pathway under a necessity criterion: removing an enabling contribution would prevent the target contribution from being realized in its claimed form. Each pathway includes enabling contributions, prior-work groundings or unmapped decisions, functional roles, and evidence-backed rationales. \textsc{SciPaths} contains 262 expert-annotated gold pathways for benchmark evaluation and 2,444 silver pathways produced in a hindsight setting for training and large-scale analysis.

%Evaluating frontier and open-weight language models, we find that current systems recover only a limited fraction of expert pathways: the best model achieves 0.189 F1 under strict semantic matching, with core methodological dependencies especially difficult to identify. Grounding improves substantially when gold enabling contributions are provided, indicating that knowing what scientific building blocks to search for is crucial for identifying the relevant prior work. These results show that scientific dependency reasoning is a challenging capability distinct from retrieving related papers or generating plausible ideas.

Evaluating frontier and open-weight language models, we find that current systems recover only a limited fraction of expert pathways: the best model achieves 0.189 F1 under strict semantic matching, with core methodological dependencies especially difficult to identify. Grounding improves substantially when gold enabling contributions are provided, indicating that knowing what scientific building blocks to search for is crucial for identifying the relevant prior work. These results show that scientific dependency reasoning is distinct from retrieving related papers or generating plausible ideas, and directly relevant to AI4Science agents: beyond proposing research directions, such systems must identify the prerequisites and prior contributions needed to make those directions feasible.

Beyond evaluation, \textsc{SciPaths} provides a training and analysis resource for modeling research trajectories as structured dependency pathways. Its annotations support studies of pathway structure, enabling roles, rationales, prior-work grounding and downstream usage. We release the silver data for training and analysis, the development set for evaluation, and the silver-construction pipeline for scaling pathway annotations to new papers, while reserving held-out test labels for benchmark evaluation.

\section{Forecasting Pathways to Scientific Discovery}
\label{sec:task}
\begin{figure*}
    \centering
    \includegraphics[width=\linewidth]{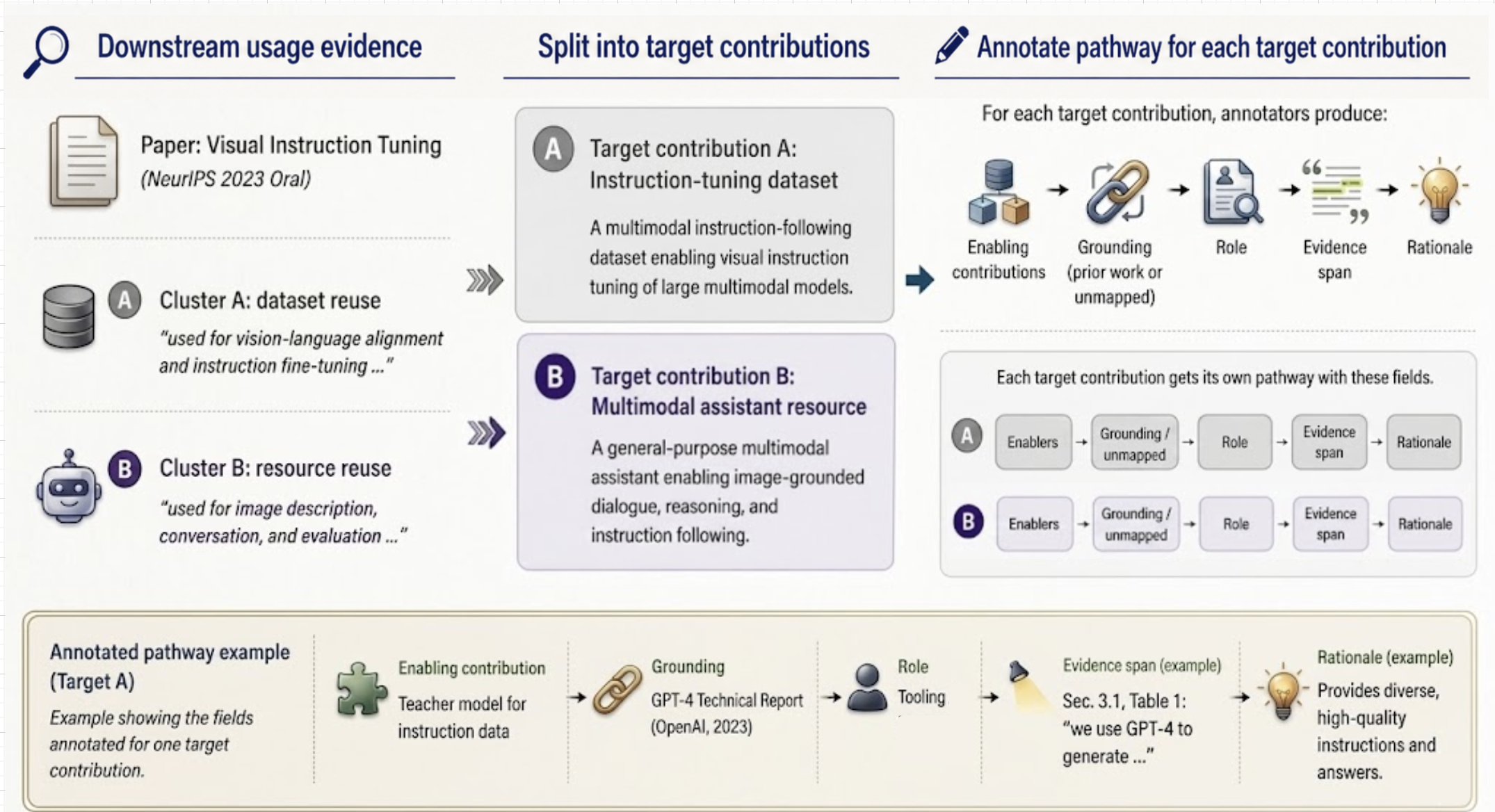}
    \caption{
Constructing \textsc{SciPaths} from downstream usage evidence. Downstream citation contexts are clustered by the contribution being reused, allowing a single paper to yield multiple target contributions. For each target contribution, expert annotators construct a separate discovery pathway containing enabling contributions, prior-work groundings or unmapped decisions, functional roles, and evidence-backed rationales. The bottom row shows one annotated pathway field example for the instruction-tuning dataset target.
}
    \label{fig:construction}
\end{figure*}
We formalize \emph{discovery pathway forecasting} as the task of identifying the \textit{enabling contributions} required to make a \textit{target contribution} feasible and grounding those in prior work when possible. A target contribution $d$ is a method, dataset, benchmark, tool, resource, or finding that subsequent work demonstrably builds upon. Let $t_d$ denote its publication time, and let $\mathcal{C}_{<t_d}$ denote the papers published before $t_d$.

Each target contribution $d$ is associated with a set of enabling contributions
\[
\mathcal{I}^*(d)=\{i_1,\ldots,i_k\},
\]
where each $i_j$ is a functional component required to realize $d$. We treat $i_j$ as necessary if removing it would prevent $d$ from being realized in its claimed form.

Each enabling contribution may be grounded in zero, one, or more representative prior papers. We define a grounding function
\[
\phi: \mathcal{I}^*(d) \rightarrow 2^{\mathcal{C}_{<t_d}},
\]
where \(2^{\mathcal{C}_{<t_d}}\) denotes the power set of the pre-target corpus, so \(\phi(i_j) \subseteq \mathcal{C}_{<t_d}\) is the set of prior papers that realize \(i_j\). If no prior paper realizes \(i_j\), then \(\phi(i_j)=\emptyset\) and the contribution is marked as \emph{unmapped}. 

A discovery pathway for target contribution $d$ is the annotated object
\[
\mathcal{P}(d)=(d,\mathcal{I}^*(d),\phi,\rho,r),
\]
where $\mathcal{I}^*(d)$ is the set of enabling contributions, and $\phi$, $\rho$, and $r$ are annotation maps over those contributions: $\phi(i_j)$ gives the prior-work groundings for $i_j$, $\rho(i_j)$ gives its functional role, and $r(i_j)$ gives its rationale. Thus, a pathway records which contributions are enabling, what role each plays, which prior work, if any, realizes it, and why it is necessary for the target contribution.
%where $\mathcal{I}^*(d)$ is the set of enabling contributions, $\phi$ maps each enabling contribution to prior-work groundings, $\rho$ assigns a functional role to each enabling contribution, and $r$ assigns a rationale to each enabling contribution. Thus, a pathway records which contributions are enabling, what role each plays, which prior work, if any, realizes it, and why it is necessary for the target contribution.
Evidence spans are included in the released annotations to support these decisions, but are not part of the core prediction target. Since multiple decompositions may be valid, $\mathcal{P}(d)$ represents one plausible, evidence-grounded pathway rather than the only possible account of how the target contribution was realized.

Given $d$ and $\mathcal{C}_{<t_d}$, models infer $\mathcal{P}(d)$ in two stages: Task A, \emph{enabling-contribution generation}, predicts $\hat{\mathcal{I}}(d)$ with roles and rationales; Task B, \emph{prior-work grounding}, maps each predicted contribution to prior work or marks it as unmapped. Unlike citation prediction, the objective is not to recover the target paper's reference list, but to identify the components required for a target contribution and the prior work, if any, that functionally realizes them.

\section{ \textsc{SciPaths} Benchmark Construction}
\label{sec:scipaths}
We now describe how \textsc{SciPaths} constructs discovery pathways from downstream usage evidence. Each instance starts from a reused target contribution and is annotated according to the pathway schema in Section~\ref{sec:task}. Figure~\ref{fig:construction} summarizes the construction process.

\subsection{Selecting Target Contributions from Downstream Reuse}
\label{sec:contribution_identification}

We construct \textsc{SciPaths} from machine learning and natural language processing papers published at NeurIPS, ICML, ACL, and EMNLP from 2023--2025. Our goal is not to annotate every contribution in each paper, but to select target contributions that later work demonstrably builds upon. These selected contributions become the target inputs for \textsc{SciPaths}: each is treated as a contribution to be realized, and expert annotators construct a pathway for that target. Downstream reuse provides a practical selection signal: contexts indicating functional dependence suggest that the reused contribution is a suitable target for pathway annotation.

Because citations serve many functions, including background, comparison, and motivation, we first filter for citation contexts that indicate functional reuse. Following \citet{shui2024}, we apply a citation-intent classifier trained on ACL-ARC \citep{jurgens2018} to identify \textsc{Uses} and \textsc{Extends} contexts, corresponding to methodological, conceptual, or resource-level reuse. We then apply LLM-based verification as a high-precision second pass to remove false positives, such as contexts that only mention that another work uses the cited paper rather than showing that the citing paper itself uses or extends it. From each verified reuse context, we extract a concise contribution description of what is reused.

We embed these contribution descriptions with a sentence encoder \citep{reimers2019} and cluster them by semantic similarity to consolidate repeated uses across independent citing papers. Each cluster yields a candidate target contribution, together with downstream usage evidence, for expert validation and pathway annotation.

\subsection{Expert Pathway Annotation}
\label{sec:annotation_setup}

Figure~\ref{fig:construction} illustrates the annotation workflow. Starting from downstream reuse evidence, annotators identify which reused contributions should become target contributions. A single paper can yield multiple targets: in the example, downstream contexts for \textit{Visual Instruction Tuning} separate into an instruction-tuning dataset target and a multimodal assistant resource target. Annotators then construct a separate pathway for each target, recording enabling contributions, prior-work groundings or unmapped decisions, roles, evidence spans, and rationales.

For each validated target, the protocol has four steps. First, annotators rewrite the target contribution at the appropriate abstraction level, capturing the \textit{object}, \textit{key property}, and \textit{what is enabled}. Second, they identify the essential enabling contributions under a constructive necessity criterion: a contribution is included only if removing it would prevent the target from being realized in its claimed form. Third, they ground each enabling contribution in representative prior work when possible, or mark it as unmapped. Finally, they assign functional roles and record evidence spans and rationales explaining necessity and grounding decisions. Role definitions are provided in Appendix~\ref{app:roles}.

We developed the protocol through pilot studies in which four annotators labeled a shared set of five papers, iteratively refining decomposition criteria, cluster-splitting rules, the interface, role definitions, and guidelines. Gold annotations were produced by five expert machine learning researchers. Annotating a single pathway typically takes 45--60 minutes. %without assistance.
On an 8-paper pilot set, annotators agreed on target selection for all papers, yielding 10 shared target contributions. Enabling-contribution decomposition achieved 74.1\% macro-averaged pairwise agreement after aligning semantically equivalent contributions, and grounding agreement over matched contributions was 90.3\%.

Given the time needed for the annotation of a pathway, we also examined optional LLM-assisted review during the pilot. Two annotators used different LLMs as auxiliary review tools, while the remaining annotators did not use LLM assistance. Assisted annotators first read the target paper and drafted their own decomposition, then used LLMs to clarify paper details, check for omitted candidate contributions, and help phrase rationales or interface responses. Final inclusion, grounding, evidence, and rationale decisions always remained with the expert annotators. Agreement was similar across assisted and unassisted annotator pairs, so optional LLM-assisted review was allowed in the final workflow. Full guidelines and protocol details are provided in Appendix~\ref{app:annotation}.
\subsection{Scaling with Silver Pathways}
\label{sec:silver}

In addition to the expert-annotated benchmark, we construct silver pathways for training and large-scale analysis. Silver pathways follow the gold schema but are produced automatically in a hindsight setting using the target paper and downstream evidence clusters. The pipeline mirrors the expert protocol: a frontier LLM (Gemini 3.1 Pro) validates downstream usage evidence, expresses the target at the appropriate abstraction level, identifies enabling contributions, grounds each in prior work or marks it as unmapped, and records roles, evidence spans, and rationales.

We prompt the model with the annotation protocol and detailed few-shot examples covering target splitting, decomposition, grounding decisions, excluded non-enabling candidates, evidence spans, and rationales. The pipeline generates multiple candidate pathways and uses a critic to select among them based on necessity, sufficiency, functional relevance, and evidence quality. On the development set, silver pathways achieve roughly 60\% F1 for enabling-contribution decomposition under the strict judge used in the main benchmark, and 68.5\% under a more permissive high-recall judge. Details on silver annotation and validation are in Appendix~\ref{app:silver}. %This is an 8.6-point improvement over the initial silver-generation baseline; gains in target splitting and grounding are smaller, as these already had higher baseline agreement.

\section{Experimental Setup}
\label{sec:experiments}

\textsc{SciPaths} comprises two tasks. Task A tests whether models can infer the enabling contributions required for a target contribution. Task B tests whether prior work realizing those contributions can be identified under a fixed literature-search budget.

\subsection{Data and Splits}
\textsc{SciPaths} contains 262 expert-annotated gold pathways and 2,444 silver pathways. We split gold pathways at the target-paper level into 50 development claims and 212 held-out test claims. The development set is used for prompt design, judge calibration, and model selection; all main results are reported on the held-out test set. We release the development set and silver pathways \footnote{\url{https://github.com/ericchamoun/scipaths}}, while reserving held-out test labels for benchmark evaluation.

\subsection{Task A: Enabling Contribution Generation}

Given a target contribution, models generate a set of enabling contributions, each with a functional description, role, and rationale. Our main setting provides only the target contribution, with no additional paper context. We also evaluate diagnostic variants to identify bottlenecks: citation-context evidence, target-paper Related Work, and few-shot examples test whether models are limited by missing context, unfamiliar output structure, or the underlying pathway reasoning itself.

\paragraph{Evaluation.}
We evaluate predicted contribution sets using semantic one-to-one matching. For each target contribution, an LLM judge labels whether each predicted--gold pair expresses the same functional requirement. For official metrics, only full semantic matches count as positive; partial or related matches are retained for diagnostic analysis but do not contribute to precision or recall. We then compute a maximum bipartite matching over matched pairs using the Hungarian algorithm, so that each predicted and gold contribution can be matched at most once. This prevents broad predictions from receiving credit for multiple distinct gold contributions. We report precision, recall, F1, and the average number of predicted contributions per target.

We use Gemini~3.1~Pro as the primary semantic matching judge. We selected it using a 60-pair human validation set stratified across clear matches, non-matches, partial matches, and judge disagreements. Gemini Flash was higher-recall but lower-precision, while Gemini~3.1~Pro was stricter and higher-precision (see Appendix~\ref{app:judge_validation} for details). Because false positives inflate pathway recovery under our strict metric, we use Gemini~3.1~Pro for primary results and report Flash robustness in Appendix~\ref{app:gemini_flash}. 

\subsection{Task B: Prior-Work Grounding}
Task B evaluates whether systems can identify prior papers that realize the enabling contributions in a target contribution's pathway. We compare four evidence conditions. All receive the target contribution claim: (1) \emph{claim-only} receives no additional information; (2) \emph{gold-contribution} receives the expert enabling contributions, giving an oracle decomposition; (3) \emph{predicted-contribution} receives all enabling contributions generated by a Task A model, representing the end-to-end setting; and (4) \emph{matched-predicted} receives only predicted contributions that semantically match gold contributions, isolating grounding when decomposition succeeds.

All conditions use the same fixed-budget Semantic Scholar pipeline. For each target contribution and evidence condition, the system generates five queries, retains the top 20 results per query, merges and deduplicates candidates, removes the target paper and papers published after $t_d$, ranks the remaining papers, and evaluates the top-$K$ results for $K \in \{5,10\}$.
\begin{table*}[t]
\caption{
Task A: Enabling-contribution generation in the main claim-only setting, evaluated on the held-out test set with Gemini~3.1~Pro as the semantic matching judge. Metrics use strict semantic one-to-one matching against expert annotations.
}
\centering
\small
\begin{tabular}{lrrrr}
\toprule
Model & Recall & Precision & F1 & Pred./target \\
\midrule
Gemini 3.1 Pro & 0.246 & 0.162 & 0.189 & 5.18 \\
Gemini Flash & 0.243 & 0.135 & 0.168 & 5.95 \\
GPT-5.4 & 0.217 & 0.123 & 0.152 & 5.96 \\
GPT-4.1 & 0.156 & 0.089 & 0.111 & 6.00 \\
GPT-5 Mini & 0.161 & 0.062 & 0.088 & 9.49 \\
Gemma-4-E4B-it & 0.069 & 0.058 & 0.061 & 4.39 \\
Qwen3-4B-Instruct & 0.070 & 0.056 & 0.060 & 4.43 \\
GPT-4o & 0.085 & 0.044 & 0.056 & 6.36 \\
Llama-3-8B-Instruct & 0.049 & 0.038 & 0.041 & 4.12 \\
Llama-3.1-8B-Instruct & 0.045 & 0.023 & 0.029 & 6.18 \\
\bottomrule
\end{tabular}

\label{tab:task-a-main}
\end{table*}

We also run an enabling-contribution-level grounding diagnostic. Given a target contribution claim, one gold enabling contribution, and its role, the model must either identify an acceptable prior paper that realizes the contribution or mark it as unmapped. This isolates grounding decisions from contribution-generation errors, and tests whether models can balance selecting prior work against abstaining when no clean grounding exists.

\paragraph{Evaluation.}
We report paper-level precision@K, recall@K, and F1@K, where a retrieved paper is correct if it matches an acceptable gold grounding. We also report enabling-contribution coverage@K: the fraction of mapped gold enabling contributions for which at least one acceptable grounding paper appears in the top-$K$ list. Coverage complements paper-level recall because some enabling contributions have multiple acceptable groundings while others have only one. Candidate coverage computes the same measure over the full retrieved candidate pool before reranking, separating retrieval failures from ranking failures. For the enabling-contribution-level diagnostic, we report mapped accuracy, unmapped accuracy, precision, recall, and recall conditioned on retrieval, where Recall$\,|$retrieved measures grounding recall among cases for which at least one acceptable grounding appears in the retrieved candidate pool.

\section{Results}
\label{sec:results}
We report main results on the 212 held-out test examples.

\subsection{Task A: Enabling-Contribution Generation}

\paragraph{Current models recover only a small fraction of expert pathways.}
Table~\ref{tab:task-a-main} reports Task A in the main claim-only setting. Under strict one-to-one semantic matching, the best model, Gemini~3.1~Pro, reaches only 0.189 F1 and 0.246 recall. Gemini Flash and GPT-5.4 follow at 0.168 and 0.152 F1, while open-weight baselines remain near or below 0.06 F1. This shows that expert pathway recovery remains difficult even when the target contribution is given.

\paragraph{Additional evidence helps, but does not close the gap.}
Appendix~\ref{app:task_a_context} reports prompting and input variants. Citation-context evidence and target-paper Related Work improve over the claim-only setting for all representative models. GPT-5.4 improves from 0.152 F1 to 0.200 with citation contexts and 0.212 with Related Work; Gemini~3.1~Pro improves from 0.189 to 0.217 with citation contexts. These gains show that context helps, but even richer inputs remain far below expert pathway recovery.

\paragraph{Silver supervision improves task alignment.}
Fine-tuning Gemma-4-E4B-it on silver pathways improves claim-only performance from 0.061 to 0.101 F1 (Appendix~\ref{app:task_a_context}). This suggests that silver data provides useful supervision for the pathway schema and expected contribution granularity, though the fine-tuned model remains well below frontier systems.

\paragraph{Overgeneration does not solve decomposition.}
GPT-5 Mini predicts the most contributions per target (9.49 on average), but its recall remains only 0.161. This suggests that models cannot recover pathways by trying many plausible prerequisites; they must infer which functional requirements are actually necessary for the target contribution.

\paragraph{Recency and rationales support the dependency-reasoning interpretation.}
Year-wise results show no consistent older-is-easier pattern, weakening a simple memorization explanation. In a rationale-quality diagnostic, Gemini~3.1~Pro matches the gold necessity rationale for 75.1\% of already matched contributions, suggesting that successful predictions often capture more than surface overlap (Appendix~\ref{app:task_a}).

\begin{figure*}[t]
    \centering
    \hspace{.7cm}
    \begin{minipage}[t]{0.42\linewidth}
        \centering
        \includegraphics[width=\linewidth]{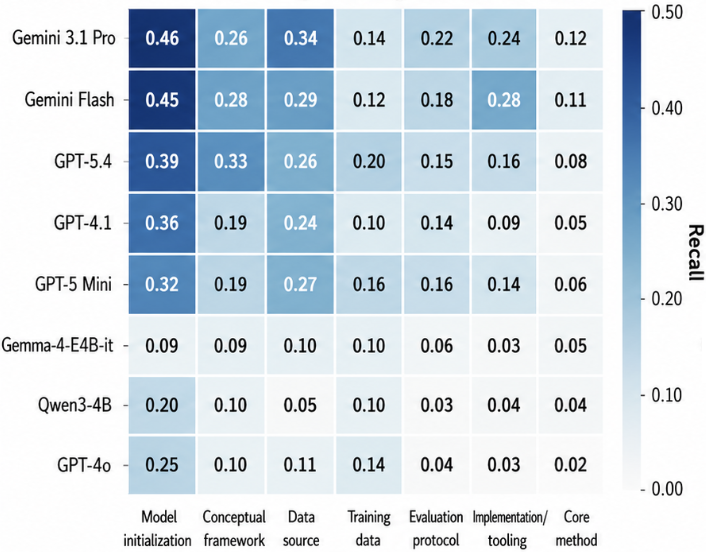}\\
        {\small (a) Recall by enabling-contribution role}
    \end{minipage}
    \hspace{1cm}
    \begin{minipage}[t]{0.38\linewidth}
        \centering
        \includegraphics[width=\linewidth]{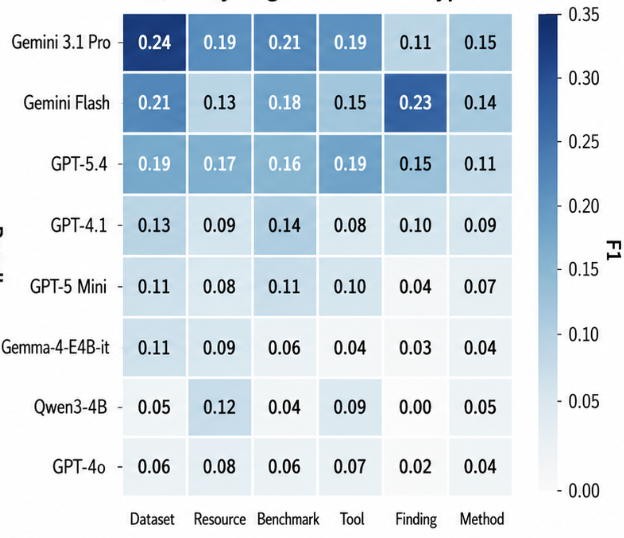}\\
        {\small (b) F1 by target-contribution type}
    \end{minipage}
    \caption{
Task A diagnostic breakdown under the Gemini~3.1~Pro judge. 
Left: recall by enabling-contribution role, showing that models recover concrete dependencies such as model initializations and data sources more reliably than core methodological contributions. 
Right: F1 by target-contribution type, showing that method and finding targets are harder to decompose than datasets, benchmarks, and tools.}
    \label{fig:heatmap}
\end{figure*}

\paragraph{Core methods are the hardest enabling contributions to recover.}
Figure~\ref{fig:heatmap} breaks down Task A by enabling-contribution role and target-contribution type. Across models, concrete dependencies such as model initializations and data sources are recovered more reliably than core methodological dependencies. Gemini~3.1~Pro recalls 0.464 of model-initialization contributions and 0.337 of data-source contributions, but only 0.119 of core-method contributions; GPT-5.4 shows the same pattern, with 0.393 recall on model initializations and 0.082 on core methods. This suggests that models can often name salient resources or pretrained backbones, but struggle to infer the specific methodological mechanisms needed to realize a target contribution. For example, a model may predict a broad prerequisite such as ``asynchronous reinforcement learning'' while missing a more specific mechanism, such as a staleness-aware data-management protocol. At the target level, method contributions are also harder to decompose than datasets and benchmarks.

\begin{table*}[t]
\caption{
Task B: Prior-work grounding on the held-out test set at $K=5$ with LLM reranking.
The Task B agent generates retrieval queries and reranks candidate papers.
The decomposition source indicates where the enabling-contribution evidence comes from: no decomposition for claim-only, model-generated contributions for predicted conditions, semantically matched model outputs for matched-predicted diagnostics, and expert annotations for gold conditions.
}
\centering
\small
\setlength{\tabcolsep}{4pt}
\renewcommand{\arraystretch}{1.08}
\resizebox{.8\linewidth}{!}{%
\begin{tabular}{lllrrrrr}
\toprule
\textbf{\begin{tabular}[c]{@{}l@{}}Task B\\agent\end{tabular}}
& \textbf{Evidence condition}
& \textbf{\begin{tabular}[c]{@{}l@{}}Decomposition\\source\end{tabular}}
& \textbf{Coverage@5}
& \textbf{Recall@5}
& \textbf{Precision@5}
& \textbf{F1@5}
& \textbf{Cand. cov.} \\
\midrule
Gemini
& Claim only & \textemdash & 0.083 & 0.055 & 0.038 & 0.041 & 0.120 \\
& Predicted contributions & Gemini & 0.062 & 0.042 & 0.030 & 0.033 & 0.122 \\
& Matched predicted & Gemini & 0.089 & 0.064 & 0.042 & 0.046 & 0.153 \\
& Predicted contributions & Gemma-4 & 0.051 & 0.032 & 0.026 & 0.027 & 0.082 \\
& Matched predicted & Gemma-4 & 0.107 & 0.053 & 0.045 & 0.046 & 0.141 \\
& Gold contributions & Expert & \textbf{0.357} & \textbf{0.270} & \textbf{0.147} & \textbf{0.172} & \textbf{0.403} \\
\midrule
GPT-5.4
& Claim only & \textemdash & 0.071 & 0.049 & 0.031 & 0.034 & 0.105 \\
& Predicted contributions & GPT-5.4 & 0.055 & 0.041 & 0.027 & 0.028 & 0.067 \\
& Matched predicted & GPT-5.4 & 0.065 & 0.042 & 0.024 & 0.027 & 0.104 \\
& Gold contributions & Expert & 0.261 & 0.195 & 0.109 & 0.127 & 0.303 \\
\bottomrule
\end{tabular}%
}
\vspace{3pt}

\label{tab:task-b-main}
\end{table*}
\paragraph{Robustness to judge choice.}
Semantic matching is sensitive to judge strictness, so we also evaluate Task A with Gemini Flash (Appendix~\ref{app:gemini_flash}), a higher-recall but lower-precision judge in our human validation. Flash raises absolute scores---GPT-5.4 and Gemini~3.1~Pro reach 0.335 and 0.330 F1---but preserves the main pattern: frontier closed-source models outperform open-weight baselines, and all models remain far from complete pathway recovery.

\subsection{Task B: Prior-Work Grounding}
\paragraph{Gold enabling contributions substantially improve prior-work recovery.}
Table~\ref{tab:task-b-main} reports Task B prior-work grounding at $K=5$ with LLM reranking. Providing expert enabling contributions substantially improves recovery for both Task B agents. With the Gemini agent, enabling-contribution Coverage@5 rises from 0.083 in the claim-only condition to 0.357 with gold enabling contributions; with the GPT-5.4 agent, coverage rises from 0.071 to 0.261. Candidate coverage also increases substantially, showing that gold enabling contributions improve the retrieved candidate pool itself, not only final reranking. However, candidate coverage remains higher than top-$K$ coverage, indicating that grounding failures arise both from missing relevant papers during retrieval and from failing to rank retrieved groundings highly enough. These results support the central hypothesis that knowing what scientific building blocks to search for is crucial for identifying the prior work that realizes them.

\paragraph{Current Task A outputs do not yet improve end-to-end grounding.}
Model-predicted enabling contributions do not reliably improve over claim-only retrieval. For the Gemini agent, raw Gemini-predicted contributions reach 0.062 Coverage@5, below the claim-only score of 0.083; GPT-5.4-predicted contributions show the same pattern with the GPT-5.4 agent. Matched-predicted diagnostics, which use only predicted contributions that semantically match gold contributions, improve modestly in some cases but remain far below gold enabling contributions. This indicates that current models do not yet generate enabling contributions that are consistently useful as search targets.

Grounding also remains difficult even when gold contributions are provided. In an enabling-contribution-level diagnostic, Gemini~3.1~Pro grounds only 26.8\% of groundable gold contributions, though it correctly leaves most unmapped contributions unmapped, with 82.4\% unmapped accuracy. This suggests that both retrieving the right prior study and deciding when no clean prior grounding exists remain challenging. Full deterministic-ranking, $K=10$, and contribution-level grounding results are reported in Appendix~\ref{app:task_b}.

Overall, Task B shows that the distinction between retrieving papers and identifying enabling contributions matters. Prior-work recovery improves substantially when gold enabling contributions are provided, but current model-generated contributions do not yet improve end-to-end grounding over direct claim retrieval. This suggests that a major bottleneck is not merely ranking papers, but identifying the right scientific building blocks to search for.
\section{Related Work}
\label{sec:related}
\textsc{SciPaths} connects AI4Science, metascience, and scientific forecasting. AI4Science systems support literature analysis, hypothesis generation, experiment design, and idea evaluation \citep{reddy2024,boiko2023,cimon,tomczak}, while metascience studies how knowledge emerges, recombines, and propagates through the literature \citep{fortunato2018,uzzi2013,wu2019,chen,zhu2022,xiang2026}. Closest to our work are scientific forecasting and citation-centered benchmarks: \textsc{PreScience} \citep{prescience} predicts key prior references at the paper level, citation-intent work studies how citation contexts signal reuse \citep{jurgens2018,shui2024}, and GIANTS \citep{giants} generates downstream insights from known parent papers. \textsc{SciPaths} targets the complementary dependency-identification step: determining which enabling contributions are required for a target contribution and which prior work, if any, realizes each one.

\section{Discussion and Limitations}
\label{sec:discussion}
\textsc{SciPaths} evaluates a capability that sits between finding relevant prior work and generating new research ideas. Its contribution-level framing reveals failures that paper-level retrieval or idea-generation evaluations may miss: a model may retrieve a relevant paper, propose a plausible idea, or identify a broadly related prerequisite while still missing the specific functional component needed to realize the target contribution. Our diagnostics show this most clearly for core methodological dependencies, which are much harder to recover than nameable resources such as data sources or model initializations. This suggests that scientific forecasting needs models that represent research trajectories as structured dependency pathways, not only as sets of papers or candidate ideas.

More broadly, \textsc{SciPaths} evaluates whether AI4Science systems can reason backward from a desired target contribution to the scientific building blocks that would make it feasible: what must already exist, what remains unmapped, and what enabling contributions would need to be developed next. Our results caution against treating current language models as standalone scientific planners. Even for machine learning and natural language processing papers that may appear in pretraining data, models struggle to reconstruct the intermediate building blocks that made a target contribution feasible.

\paragraph{Limitations.}
\textsc{SciPaths} captures observed, evidence-grounded pathways rather than a unique account of how a target contribution was realized. Multiple decompositions may be valid, and experts may disagree about granularity or necessity despite our guidelines, rationales, and semantic matching protocol. Task A relies on an LLM judge for semantic matching; although we validate the judge against expert annotations and report higher-recall robustness results, judgment errors may affect absolute scores. Task B depends on Semantic Scholar search and metadata, so failures can reflect search limitations or incomplete metadata. Finally, the benchmark focuses on machine learning and natural language processing papers; extending it to other fields may require adapting role definitions and annotation guidelines.

\section{Conclusion}
\label{sec:conclusion}
We introduced \textsc{SciPaths}, a benchmark for discovery pathway forecasting. Unlike paper-level citation benchmarks, \textsc{SciPaths} represents target contributions as pathways of enabling contributions, prior-work groundings when available, and unmapped decisions otherwise. Across frontier and open-weight language models, we find that current systems recover only a small fraction of expert pathways under strict semantic matching, with core methodological dependencies especially difficult to identify. Prior-work grounding improves substantially when gold enabling contributions are provided, but end-to-end performance remains limited by decomposition quality. We hope \textsc{SciPaths} supports future work on models that reason about the contribution-level dependency structure of scientific progress.

 \section*{Acknowledgments}

This research was developed with funding
from the Defense Advanced Research Projects
Agency’s (DARPA) SciFy program (Agreement
No. HR00112520300). Eric Chamoun is supported by an EPSRC-funded
studentship.

% In the unusual situation where you want a paper to appear in the
% references without citing it in the main text, use \nocite
%\nocite{langley00}

\bibliography{main}

@article{
fortunato2018,
author = {Santo Fortunato  and Carl T. Bergstrom  and Katy Börner  and James A. Evans  and Dirk Helbing  and Staša Milojević  and Alexander M. Petersen  and Filippo Radicchi  and Roberta Sinatra  and Brian Uzzi  and Alessandro Vespignani  and Ludo Waltman  and Dashun Wang  and Albert-László Barabási },
title = {Science of science},
journal = {Science},
volume = {359},
number = {6379},
pages = {eaao0185},
year = {2018},
doi = {10.1126/science.aao0185},
URL = {https://www.science.org/doi/abs/10.1126/science.aao0185},
eprint = {https://www.science.org/doi/pdf/10.1126/science.aao0185}}

@article{
uzzi2013,
author = {Brian Uzzi  and Satyam Mukherjee  and Michael Stringer  and Ben Jones },
title = {Atypical Combinations and Scientific Impact},
journal = {Science},
volume = {342},
number = {6157},
pages = {468-472},
year = {2013},
doi = {10.1126/science.1240474},
URL = {https://www.science.org/doi/abs/10.1126/science.1240474},
eprint = {https://www.science.org/doi/pdf/10.1126/science.1240474}}

@inproceedings{reddy2024,
  author={Chandan K. Reddy and Parshin Shojaee},
  title={Towards Scientific Discovery with Generative AI: Progress, Opportunities, and Challenges},
  year={2025},
  cdate={1735689600000},
  pages={28601-28609},
  url={https://doi.org/10.1609/aaai.v39i27.35084},
  booktitle={AAAI}
}

@article{boiko2023,
  title={Autonomous chemical research with large language models},
  author={Daniil A. Boiko and Robert MacKnight and Ben Kline and Gabe Gomes},
  journal={Nature},
  year={2023},
  volume={624},
  pages={570 - 578},
  url={https://api.semanticscholar.org/CorpusID:266432059}
}

@article{wu2019,
  title={Large teams develop and small teams disrupt science and technology},
  author={Lingfei Wu and Dashun Wang and James A. Evans},
  journal={Nature},
  year={2019},
  volume={566},
  pages={378 - 382},
  url={https://api.semanticscholar.org/CorpusID:61156556}
}

@inproceedings{cimon,
    title = "{S}ci{MON}: Scientific Inspiration Machines Optimized for Novelty",
    author = "Wang, Qingyun  and
      Downey, Doug  and
      Ji, Heng  and
      Hope, Tom",
    editor = "Ku, Lun-Wei  and
      Martins, Andre  and
      Srikumar, Vivek",
    booktitle = "Proceedings of the 62nd Annual Meeting of the Association for Computational Linguistics (Volume 1: Long Papers)",
    month = aug,
    year = "2024",
    address = "Bangkok, Thailand",
    publisher = "Association for Computational Linguistics",
    url = "https://aclanthology.org/2024.acl-long.18/",
    doi = "10.18653/v1/2024.acl-long.18",
    pages = "279--299"
}

@misc{prescience,
      title={PreScience: A Benchmark for Forecasting Scientific Contributions}, 
      author={Anirudh Ajith and Amanpreet Singh and Jay DeYoung and Nadav Kunievsky and Austin C. Kozlowski and Oyvind Tafjord and James Evans and Daniel S. Weld and Tom Hope and Doug Downey},
      year={2026},
      eprint={2602.20459},
      archivePrefix={arXiv},
      primaryClass={cs.AI},
      url={https://arxiv.org/abs/2602.20459}, 
}

@article{tomczak,
author = {Tomczak, Maciej and Park, Yang and Hsu, Chia‐Wei and Brown, Payden and Massa, Dario and Sankowski, Piotr and Li, Ju and Papanikolaou, Stefanos},
year = {2025},
month = {09},
pages = {},
title = {Forecasting Research Trends Using Knowledge Graphs and Large Language Models},
volume = {8},
journal = {Advanced Intelligent Systems},
doi = {10.1002/aisy.202401124}
}

@misc{chen,
      title={Structuring Scientific Innovation: A Framework for Modeling and Discovering Impactful Knowledge Combinations}, 
      author={Junlan Chen and Kexin Zhang and Daifeng Li and Yangyang Feng and Yuxuan Zhang and Bowen Deng},
      year={2025},
      eprint={2503.18865},
      archivePrefix={arXiv},
      primaryClass={cs.AI},
      url={https://arxiv.org/abs/2503.18865}, 
}

@inproceedings{zhu2022,
    title = "Predicting Prerequisite Relations for Unseen Concepts",
    author = "Zhu, Yaxin  and
      Zamani, Hamed",
    editor = "Goldberg, Yoav  and
      Kozareva, Zornitsa  and
      Zhang, Yue",
    booktitle = "Proceedings of the 2022 Conference on Empirical Methods in Natural Language Processing",
    month = dec,
    year = "2022",
    address = "Abu Dhabi, United Arab Emirates",
    publisher = "Association for Computational Linguistics",
    url = "https://aclanthology.org/2022.emnlp-main.585/",
    doi = "10.18653/v1/2022.emnlp-main.585",
    pages = "8542--8548"
}

@article{xiang2026,
title = {Knowledge precedence networks: Mining progression patterns of scientific discoveries beyond prerequisites},
journal = {Information Processing \& Management},
volume = {63},
number = {2, Part B},
pages = {104424},
year = {2026},
issn = {0306-4573},
doi = {https://doi.org/10.1016/j.ipm.2025.104424},
url = {https://www.sciencedirect.com/science/article/pii/S0306457325003656},
author = {Shibing Xiang and Bing Liu and Xin Jiang and Zhengan Huang and Yifang Ma}
}

@article{jurgens2018,
    title = "Measuring the Evolution of a Scientific Field through Citation Frames",
    author = "Jurgens, David  and
      Kumar, Srijan  and
      Hoover, Raine  and
      McFarland, Dan  and
      Jurafsky, Dan",
    editor = "Lee, Lillian  and
      Johnson, Mark  and
      Toutanova, Kristina  and
      Roark, Brian",
    journal = "Transactions of the Association for Computational Linguistics",
    volume = "6",
    year = "2018",
    address = "Cambridge, MA",
    publisher = "MIT Press",
    url = "https://aclanthology.org/Q18-1028/",
    doi = "10.1162/tacl_a_00028",
    pages = "391--406"
}

@inproceedings{shui2024,
    title = "Fine-Tuning Language Models on Multiple Datasets for Citation Intention Classification",
    author = "Shui, Zeren  and
      Karypis, Petros  and
      Karls, Daniel S.  and
      Wen, Mingjian  and
      Manchanda, Saurav  and
      Tadmor, Ellad B.  and
      Karypis, George",
    editor = "Al-Onaizan, Yaser  and
      Bansal, Mohit  and
      Chen, Yun-Nung",
    booktitle = "Findings of the Association for Computational Linguistics: EMNLP 2024",
    month = nov,
    year = "2024",
    address = "Miami, Florida, USA",
    publisher = "Association for Computational Linguistics",
    url = "https://aclanthology.org/2024.findings-emnlp.974/",
    doi = "10.18653/v1/2024.findings-emnlp.974",
    pages = "16718--16732"
}

@inproceedings{reimers2019,
    title = "Sentence-{BERT}: Sentence Embeddings using {S}iamese {BERT}-Networks",
    author = "Reimers, Nils  and
      Gurevych, Iryna",
    editor = "Inui, Kentaro  and
      Jiang, Jing  and
      Ng, Vincent  and
      Wan, Xiaojun",
    booktitle = "Proceedings of the 2019 Conference on Empirical Methods in Natural Language Processing and the 9th International Joint Conference on Natural Language Processing (EMNLP-IJCNLP)",
    month = nov,
    year = "2019",
    address = "Hong Kong, China",
    publisher = "Association for Computational Linguistics",
    url = "https://aclanthology.org/D19-1410/",
    doi = "10.18653/v1/D19-1410",
    pages = "3982--3992"
}

@misc{giants,
      title={GIANTS: Generative Insight Anticipation from Scientific Literature}, 
      author={Joy He-Yueya and Anikait Singh and Ge Gao and Michael Y. Li and Sherry Yang and Chelsea Finn and Emma Brunskill and Noah D. Goodman},
      year={2026},
      eprint={2604.09793},
      archivePrefix={arXiv},
      primaryClass={cs.CL},
      url={https://arxiv.org/abs/2604.09793}, 
}

@inproceedings{visual-it,
 author = {Liu, Haotian and Li, Chunyuan and Wu, Qingyang and Lee, Yong Jae},
 booktitle = {Advances in Neural Information Processing Systems},
 editor = {A. Oh and T. Naumann and A. Globerson and K. Saenko and M. Hardt and S. Levine},
 pages = {34892--34916},
 publisher = {Curran Associates, Inc.},
 title = {Visual Instruction Tuning},
 url = {https://proceedings.neurips.cc/paper_files/paper/2023/file/6dcf277ea32ce3288914faf369fe6de0-\\Paper-Conference.pdf},
 volume = {36},
 year = {2023}
}
\bibliographystyle{icml2026}

%%%%%%%%%%%%%%%%%%%%%%%%%%%%%%%%%%%%%%%%%%%%%%%%%%%%%%%%%%%%%%%%%%%%%%%%%%%%%%%
%%%%%%%%%%%%%%%%%%%%%%%%%%%%%%%%%%%%%%%%%%%%%%%%%%%%%%%%%%%%%%%%%%%%%%%%%%%%%%%
% APPENDIX
%%%%%%%%%%%%%%%%%%%%%%%%%%%%%%%%%%%%%%%%%%%%%%%%%%%%%%%%%%%%%%%%%%%%%%%%%%%%%%%
%%%%%%%%%%%%%%%%%%%%%%%%%%%%%%%%%%%%%%%%%%%%%%%%%%%%%%%%%%%%%%%%%%%%%%%%%%%%%%%
\newpage
\appendix
\onecolumn

\section{Annotation Details}
\label{app:annotation}

The annotation guidelines below summarize the annotator-facing protocol used during data collection. 

\subsection{Overview}

The goal of \textsc{SciPaths} annotation is to identify, for each selected target contribution, the enabling contributions required to realize it and the prior work, if any, that realizes each enabling contribution. The annotation procedure has two substantive phases:
\begin{enumerate}
    \item \textbf{Target contribution assessment}: validate downstream reuse evidence and rewrite the target contribution at the appropriate level of abstraction.
    \item \textbf{Enabling-contribution annotation}: decompose the target contribution into necessary enabling contributions, ground each contribution in representative prior work when available or mark it as unmapped, assign roles, and justify each dependency.
\end{enumerate}

The guiding counterfactual is:
\begin{quote}
\emph{If I had to realize this target contribution tomorrow, what enabling contributions would I still need?}
\end{quote}

This shifts annotation from citation recovery to enabling-contribution recovery. Annotators are not asked to list all relevant references, but to identify the functional requirements without which the target contribution could not be realized in its claimed form.

\subsection{Phase 1: Target Contribution Assessment}
\label{app:phase1}

\paragraph{Goal.}
The goal of Phase 1 is to determine whether a paper contains one or more valid target contributions supported by downstream reuse evidence, and to rewrite each target contribution at the correct level of abstraction.

A valid target contribution is:
\begin{quote}
\emph{A contribution, such as a method, dataset, benchmark, tool, resource, or finding, that subsequent work depends on to build, evaluate, or extend its own work.}
\end{quote}

This phase focuses on functional dependence, not popularity or citation frequency.

\paragraph{Decision procedure.}
Annotators inspect the candidate contribution and downstream usage clusters, verify whether later work functionally depends on the contribution, and then decide whether the contribution should be retained. Strong evidence includes direct reuse, training dependence, evaluation adoption, extension, adaptation, or other forms of functional dependence. Background citations, comparison-only usage, weak one-off mentions, or hallucinated cluster summaries are not sufficient.

A single paper may contain multiple target contributions. Annotators split contributions when a paper introduces distinct reusable outputs, such as a model and a benchmark, that enable different downstream uses and would require different enabling contributions. Annotators do not bundle multiple target contributions into a single claim.

\paragraph{Rewriting target contribution claims.}
Each rewritten target contribution should preserve:
\[
\text{[object]} + \text{[key property]} + \text{[what it enables]}.
\]
A valid rewritten claim should be:
\begin{itemize}
    \item \textbf{Atomic}: describes one contribution only;
    \item \textbf{Abstracted}: avoids paper-specific names when possible;
    \item \textbf{Functional}: states what the contribution does;
    \item \textbf{Causal}: specifies what the contribution enables;
    \item \textbf{Decomposable}: can be broken into enabling contributions in Phase 2.
\end{itemize}

For example:
\begin{quote}
\emph{Benchmark: A multi-turn dialogue sentiment reasoning benchmark, enabling evaluation of cross-utterance opinion and sentiment understanding.}
\end{quote}
is preferred over:
\begin{quote}
\emph{The DiaASQ benchmark and dataset.}
\end{quote}

Common failure modes include name-based claims, bundled claims, vague claims such as ``improves performance,'' motivational claims, and implementation-level details.

\subsection{Phase 2: Enabling-Contribution Annotation}
\label{app:phase2}

\paragraph{Goal.}
The goal of Phase 2 is to identify the enabling contributions required to realize the validated target contribution and to ground each enabling contribution in prior work when available. This phase is not about selecting all relevant citations. It reconstructs the pathway through necessary functional components:
\[
\text{target contribution} \rightarrow \text{enabling contributions} \rightarrow \text{prior-work grounding or unmapped}.
\]

\paragraph{Core reasoning principles.}
Annotators follow three principles:
\begin{enumerate}
    \item \textbf{Necessity}: each enabling contribution must be something without which the target contribution could not be realized in its claimed form.
    \item \textbf{Functional abstraction}: enabling contributions should be expressed as capabilities, substrates, formulations, objectives, upstream resources, or mechanisms, not as paper sections, hyperparameters, or arbitrary citations.
    \item \textbf{Evidence support}: evidence spans must come from the target paper and directly support the contribution--role--grounding decision.
\end{enumerate}

\paragraph{Valid enabling contributions.}
A valid enabling contribution is a necessary functional requirement or upstream substrate for the target contribution. Common types include task formulations, conceptual paradigms, source datasets, training data, model initializations, objectives, representations, source websites or raw corpora, implementation resources, and evaluation protocols when central to the target contribution.

Good examples include:
\begin{itemize}
    \item federated learning training and aggregation protocol for client--server PLM tuning;
    \item semantically aligned visual encoder for image understanding;
    \item upstream Turkish Wikipedia NER substrate for re-annotation;
    \item cross-utterance quadruple composition in dialogue.
\end{itemize}

Bad examples include:
\begin{itemize}
    \item training for three epochs;
    \item stronger baseline models;
    \item methods section;
    \item evaluation on benchmark X when the benchmark is not part of the target contribution.
\end{itemize}

\paragraph{Grounding and unmapped decisions.}
For each enabling contribution, annotators choose a canonical grounding for annotation purposes:
\begin{itemize}
    \item a representative prior study/resource, or
    \item \texttt{NONE}, meaning no single prior study or resource cleanly represents the enabling contribution.
\end{itemize}

A prior work should be selected when it directly provides, instantiates, or is reused as the enabling contribution. \texttt{NONE} should be selected when the enabling contribution is field-level, composite, paper-specific, or otherwise not attributable to a single clean prior study. This is a valid outcome: annotators should not force weak or fake groundings to avoid \texttt{NONE}. Additional valid studies/resources may be attached when several sources jointly instantiate an enabling contribution or when one canonical grounding is representative but not exhaustive.

A canonical grounding should be the cleanest representative of the enabling contribution: necessary rather than merely related, minimal rather than overly broad, and faithful to the actual role played in the target paper.

\subsection{Functional Role Definitions}
\label{app:roles}

Annotators assign one role from the approved role set:
\begin{itemize}
    \item \textbf{Core Method / Algorithm}: a prior method or algorithmic procedure that provides a necessary computational mechanism used to realize the target contribution, such as a training objective, model architecture, optimization procedure, or inference algorithm.
    \item \textbf{Conceptual Framework}: prior work that defines the task, problem formulation, representation, theoretical framework, or empirical phenomenon that the target contribution builds upon.
    \item \textbf{Data Source}: a dataset, corpus, website, or resource explicitly used as source material to construct another dataset or resource.
    \item \textbf{Training Data}: a dataset or labeled resource directly used to train, pretrain, fine-tune, or supervise a model. If a dataset is transformed, sampled, re-annotated, translated, or used to build a new dataset, annotators use \textbf{Data Source} instead.
    \item \textbf{Model Initialization}: a pretrained model or initialization essential to realizing the target contribution, such as initializing with pretrained BERT weights.
    \item \textbf{Evaluation Protocol}: a benchmark, metric, or annotation scheme directly reused and necessary to realize the target contribution. Benchmarks used only for breadth or comparison are excluded.
    \item \textbf{Implementation / Tooling}: software, infrastructure, or tooling explicitly required to implement the target contribution.
\end{itemize}

\subsection{Annotation Fields}

For each enabling contribution, annotators record:
\begin{itemize}
    \item \textbf{Enabling contribution}: the functional component needed for the target contribution.
    \item \textbf{Canonical grounding}: the representative prior study/resource or \texttt{NONE}.
    \item \textbf{Additional groundings}: optional additional valid studies/resources.
    \item \textbf{Role}: one of the approved functional roles.
    \item \textbf{Contribution}: what the selected study/resource provides.
    \item \textbf{Rationale}: why the enabling contribution is necessary and why the grounding, if any, realizes it.
    \item \textbf{Evidence span}: a sentence or short span from the target paper supporting the dependency.
\end{itemize}

A good rationale answers: what is needed, why it is needed, and why the selected prior work provides it. Evidence spans should directly support the dependency and role assignment, not merely provide background or related-work context.

\subsection{Quality Checklist}

Before finalizing an annotation, annotators check:
\begin{itemize}
    \item Is each enabling contribution truly necessary?
    \item Is it a functional requirement rather than an implementation detail?
    \item Is the selected prior study/resource the cleanest representative?
    \item Should the contribution instead be marked as \texttt{NONE}?
    \item Are additional groundings genuinely needed?
    \item Does the rationale explain necessity rather than similarity?
    \item Does the evidence span directly support the assigned role and grounding?
\end{itemize}

Common mistakes include listing citations instead of enabling contributions, choosing a study because it is famous rather than necessary, including evaluation datasets used only for comparison, forcing a grounding when \texttt{NONE} is correct, writing vague rationales, using evidence from the wrong stage of the pipeline, and ignoring direct source resources such as websites or corpora when they are explicitly used to construct a dataset.
\subsection{Inter-Annotator Agreement}
\label{app:annotation_agreement}

We measure inter-annotator agreement (IAA) for both stages of the enabling-contribution annotation: (i) identifying the necessary enabling contributions for a target contribution, and (ii) grounding those enabling contributions in prior studies or resources.

\paragraph{Enabling-contribution agreement.}
For each target contribution, we first construct an aligned enabling-contribution universe by grouping semantically equivalent annotations. Each annotator is then represented as a binary vector over this aligned universe, indicating whether they included each enabling contribution. Pairwise agreement is computed as the fraction of enabling contributions included by either annotator that were included by both annotators:
\[
\mathrm{Agreement}(a,b) =
\frac{
|\{i \in \mathcal{E} : x_{a,i}=1 \land x_{b,i}=1\}|
}{
|\{i \in \mathcal{E} : x_{a,i}=1 \lor x_{b,i}=1\}|
},
\]
where $\mathcal{E}$ is the aligned enabling-contribution universe for the target contribution and $x_{a,i}$ indicates whether annotator $a$ included enabling contribution $i$.

\paragraph{Grounding agreement.}
Grounding agreement is computed separately from enabling-contribution agreement. For each pair of annotators, we consider only enabling contributions that both annotators included. We then compare whether their selected groundings refer to the same prior study, resource, or source family. When the same source appears as canonical for one annotator and as an additional grounding for another, we count it as agreement, since the disagreement is about placement rather than provenance. We also count \texttt{NONE} as agreement when both annotators judged that no single prior study or resource cleanly represents the enabling contribution.

For enabling contributions with multiple groundings, we compute fractional source overlap when needed. For example, if two annotators agree on two of four source-level groundings for a composite enabling contribution, that contribution receives partial grounding agreement. We then average grounding agreement across the enabling contributions shared by the annotator pair.

\paragraph{Qualitative disagreement patterns.}
Most disagreements are interpretable boundary cases rather than random contradictions. Annotators usually agree on the central enabling contributions for a target contribution, but sometimes differ on whether to represent an auxiliary or paper-specific component as a separate enabling contribution. The higher grounding agreement suggests that disagreements are mostly about enabling-contribution granularity rather than source provenance. When annotators identify the same enabling contribution, they generally select the same prior study or resource as the relevant grounding. This supports the reliability of the annotation framework: the task is difficult and high-granularity, but annotators converge on the main dependency structure and largely agree on the scientific provenance of matched enabling contributions.

\subsection{LLM Usage in Annotation}
\label{app:llm-usage}

Pathway annotation requires annotators to read the target paper, inspect downstream usage evidence, identify necessary enabling contributions, ground those contributions in prior work, and write evidence-backed rationales. During protocol development, we examined whether optional LLM-assisted review could improve annotation efficiency without changing the annotation target.

In the agreement pilot, two annotators completed the task without LLM assistance, while the remaining annotators used different LLMs as auxiliary review tools. LLM-assisted annotators first read the target paper and drafted their own decomposition before consulting the model. They could then use the LLM to clarify paper details, check whether their draft omitted plausible enabling contributions, compare alternative phrasings, or help write clearer rationales and interface responses. All gold pathways were finalized by expert annotators; LLM outputs were used only as optional review aids and were never accepted without human verification.

Final annotation decisions always remained with the expert annotators. In particular, annotators made the final decisions about (i) which target contributions should be retained, (ii) which enabling contributions satisfied the necessity criterion, (iii) whether each enabling contribution should be grounded in prior work or marked as unmapped, and (iv) which evidence spans and rationales supported the decision.

We compared agreement across assisted and unassisted annotator pairs in the pilot. Agreement was similar across pairs: enabling-contribution decomposition pairwise means ranged from 69.3--78.1\%, and grounding agreement ranged from 86.7--92.7\%. The overall macro-averaged agreement was 74.1\% for enabling-contribution decomposition and 90.3\% for grounding over matched contributions. Based on these results, we allowed optional LLM-assisted review in the final workflow. In practice, LLM assistance was most useful for improving annotation efficiency, especially by helping annotators check draft decompositions and phrase rationales after substantive pathway decisions had been made. 
\section{Silver Pathway Construction}
\label{app:silver}
\begin{figure}[H]
    \centering
    \includegraphics[width=\linewidth]{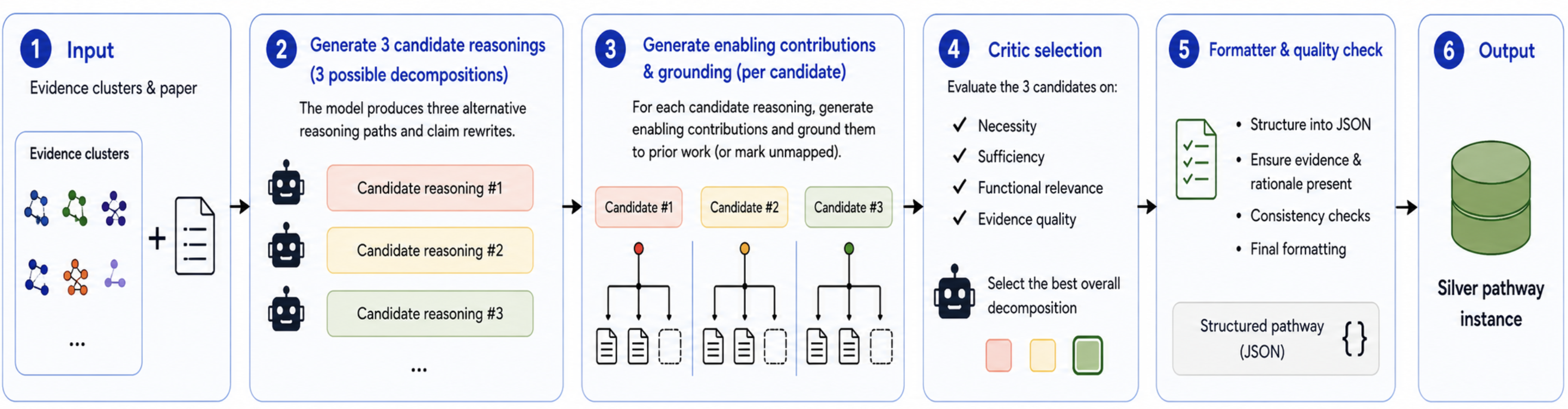}
\caption{Silver annotation pipeline overview.
}
    \label{fig:silver}
\end{figure}
We construct silver pathways to provide additional training data and to support large-scale analyses of pathway structure. Silver pathways follow the same schema as the expert gold annotations, but are produced automatically in a hindsight setting using the target paper and downstream usage evidence clusters. They are intended for training and analysis only; all benchmark evaluation is conducted on expert-annotated gold pathways. Figure~\ref{fig:silver} provides our enabling contribution decomposition pipeline.

\paragraph{Inputs.}
For each candidate target contribution, the silver pipeline receives: (i) the target paper, (ii) downstream usage clusters indicating how later work reuses the contribution, and (iii) retrieved candidate prior work from the pre-\(t_d\) corpus. The use of the target paper makes this a hindsight construction setting, analogous to the expert annotation process, rather than a model-only forecasting setting.

\paragraph{Few-shot annotation prompting.}
To align automatic annotations with the expert schema, we prompt a frontier LLM with detailed few-shot examples of complete pathway annotations. We release the prompts as part of our code. These examples demonstrate how to split bundled contributions into separate target contributions, rewrite each target at the appropriate abstraction level, identify essential enabling contributions, exclude tempting non-enabling contributions, ground contributions in prior work or mark them as unmapped, assign functional roles, and provide evidence-backed rationales. The examples also emphasize the constructive necessity criterion: an enabling contribution should be included only if removing it would prevent the target contribution from being realized in its claimed form.

\paragraph{Candidate pathway generation.}
For each candidate target contribution, the model first validates the downstream usage evidence and expresses the target contribution at the appropriate level of abstraction, preserving the object, key property, and what the contribution enables. It then generates a candidate pathway containing enabling contributions, functional roles, grounding decisions, evidence spans, and rationales. For each enabling contribution, the model either selects representative prior work from the candidate pool or marks the contribution as unmapped when no prior study cleanly realizes it.

\paragraph{Candidate selection and formatting.}
Because a single target contribution can admit multiple plausible decompositions, the pipeline generates multiple candidate pathways. A critic then selects the best candidate according to necessity, sufficiency, functional relevance, grounding quality, and evidence support. The selected pathway is converted into the \textsc{SciPaths} schema, including target contribution, enabling contributions, roles, canonical and additional groundings when available, unmapped decisions, evidence spans, and rationales. We also run consistency checks to ensure that required fields are present and that grounding decisions are compatible with the available evidence.

\paragraph{Validation against gold annotations.}
We validate silver quality on the development set by comparing silver pathways against expert gold annotations. For target-contribution agreement, we compare whether the automatic pipeline identifies the same target contribution. For enabling-contribution decomposition, we use the same semantic matching protocol as Task A. For grounding, we evaluate matched enabling contributions and check whether the silver and gold annotations select the same grounding study. Under the strict judge used for the main benchmark, silver pathways achieve approximately \(60\%\) F1 for enabling-contribution decomposition; under a more permissive high-recall judge, F1 increases to \(68.5\%\). The final silver pipeline improves enabling-contribution decomposition by \(8.6\) percentage points over the initial silver-generation baseline, with smaller gains in target splitting and grounding agreement which already had higher baseline agreement. 

These results indicate that silver pathways provide useful supervision for training and large-scale analysis, while expert gold annotations remain the standard for benchmark evaluation.
\section{Experimental Details}
\label{app:experimental_details}

\paragraph{Task A evaluation.}
Task A evaluates enabling-contribution generation against expert annotations using semantic one-to-one matching. For each predicted--gold contribution pair within a target, an LLM judge assigns a semantic label. The main metric counts only full semantic matches as correct. After pairwise judgments are obtained, we enforce a one-to-one alignment between predicted and gold contributions using maximum bipartite matching over full-match edges. We then compute precision, recall, and F1 per target and report macro averages across targets. Partial matches are excluded from the main metric and used only in a separate diagnostic.

\paragraph{Judge model.}
Unless otherwise noted, Task A results use Gemini~3.1~Pro as the semantic matching judge. We selected this judge after validating Gemini~3.1~Pro and Gemini Flash against expert human labels: Gemini Flash is more permissive and higher-recall, while Gemini~3.1~Pro is stricter and higher-precision. We therefore use Gemini~3.1~Pro for the primary benchmark results and report Gemini Flash as a robustness check.

\paragraph{Task A settings.}
The main setting (\textbf{S1}) gives the model only the target contribution. \textbf{S2} additionally provides citation-context evidence about downstream reuse, and \textbf{S3} provides target-paper Related Work context. We also report a few-shot prompting condition and a silver fine-tuning condition. These diagnostic settings test whether models improve when given richer evidence, task demonstrations, or supervised pathway data.

\paragraph{Task B evaluation.}
Task B evaluates recovery of prior work that grounds a target contribution pathway. Given a target contribution and an evidence condition, the system retrieves candidate prior papers, optionally reranks them, and is scored against expert-annotated acceptable groundings. We report contribution coverage, paper-level recall, precision, F1, and candidate-pool contribution coverage. We include both deterministic ranking and LLM reranking at budgets $K \in \{5,10\}$.

\paragraph{Contribution-level grounding diagnostic.}
To isolate grounding from decomposition, we additionally evaluate contribution-level grounding on the test set. In the oracle condition, the grounding agent receives gold enabling contributions. In the predicted condition, it receives model-predicted contributions from Task~A. This diagnostic reports mapped accuracy, grounding recall, precision, recall conditioned on retrieval, and unmapped accuracy.

\section{Task A Additional Results}
\label{app:task_a}

\subsection{Gemini Flash Judge Robustness}
\label{app:gemini_flash}

Table~\ref{tab:task_a_main_flash} reports Task~A test results using Gemini Flash as the semantic matching judge. Scores are substantially higher than under Gemini~3.1~Pro because Flash is more permissive, as shown by the judge validation in Section~\ref{app:judge_validation}. The overall pattern is stable, with frontier closed-source models outperforming open-weight baselines and all models remaining far from complete pathway recovery.

\begin{table}[t]
\caption{
Task~A enabling-contribution generation on the held-out test set in the main claim-only setting, using Gemini Flash as the semantic matching judge. Flash yields consistently higher absolute scores than Gemini~3.1~Pro, but the main qualitative conclusions remain unchanged.
}
\centering
\small
\begin{tabular}{lrrrr}
\toprule
Model & Recall & Precision & F1 & Pred./target \\
\midrule
Gemini 3.1 Pro & 0.432 & 0.282 & 0.331 & 5.22 \\
GPT-5.4 & 0.476 & 0.265 & 0.330 & 6.17 \\
Gemini Flash & 0.388 & 0.235 & 0.284 & 5.65 \\
GPT-4.1 & 0.351 & 0.215 & 0.260 & 5.79 \\
GPT-5 Mini & 0.409 & 0.163 & 0.226 & 8.82 \\
Gemma-4-E4B-it & 0.250 & 0.198 & 0.214 & 4.42 \\
GPT-4o & 0.249 & 0.140 & 0.175 & 6.26 \\
Qwen3-4B-Instruct & 0.186 & 0.144 & 0.156 & 4.30 \\
Gemma-2-2B-it & 0.141 & 0.100 & 0.113 & 4.81 \\
Llama-3-8B-Instruct & 0.129 & 0.102 & 0.111 & 4.24 \\
Llama-3.1-8B-Instruct & 0.160 & 0.085 & 0.108 & 6.36 \\
Qwen2.5-7B-Instruct & 0.110 & 0.100 & 0.102 & 3.98 \\
Llama-3.2-3B-Instruct & 0.123 & 0.086 & 0.087 & 5.45 \\
\bottomrule
\end{tabular}
\label{tab:task_a_main_flash}
\end{table}

\subsection{Judge Validation}
\label{app:judge_validation}

To select the primary semantic matching judge, we audited 60 stratified predicted--gold contribution pairs from the development set. The sample includes clear matches, clear non-matches, borderline partial matches, and cases where Gemini Flash and Gemini~3.1~Pro disagreed. Two human annotators independently assigned three-way labels: \textsc{Match}, \textsc{Partial}, and \textsc{No Match}. For validation, we collapse these labels to the official binary setting, where only \textsc{Match} counts as positive.

Table~\ref{tab:judge_validation} shows that Gemini Flash behaves as a high-recall, lower-precision judge, while Gemini~3.1~Pro is substantially stricter and higher-precision. Because the official metric is intentionally strict and false positives inflate pathway-recovery scores, we use Gemini~3.1~Pro as the primary judge and report Gemini Flash as a higher-recall robustness check.

\begin{table}[h]
\caption{
Binary judge validation on 60 stratified predicted--gold contribution pairs. Only \textsc{Match} is treated as positive; \textsc{Partial} and \textsc{No Match} are treated as negative, matching the official Task~A metric. Gemini Flash is more permissive, while Gemini~3.1~Pro is stricter and higher-precision.
}
\centering
\small
\begin{tabular}{lrrrrr}
\toprule
Comparison & $N$ & Accuracy & Precision & Recall & F1 \\
\midrule
Gemini Flash vs humans avg. & 60 & 0.800 & 0.652 & 0.977 & 0.782 \\
Gemini Pro vs humans avg. & 60 & 0.833 & 0.900 & 0.614 & 0.730 \\
Human vs human & 60 & 0.900 & 0.864 & 0.864 & 0.864 \\
\bottomrule
\end{tabular}

\label{tab:judge_validation}
\end{table}
\newpage
\subsection{Input and Prompting Variants}
\label{app:task_a_context}

Table~\ref{tab:task_a_variants_appendix} reports Task~A prompting and evidence variants on the held-out test set under Gemini~3.1~Pro judging. Adding citation context or Related Work context improves over the claim-only baseline, though the relative gains are model-dependent. Few-shot prompting generally improves precision and often improves F1, but does not consistently outperform richer context variants. Overall, these results suggest that models benefit from additional evidence and examples, but still struggle to infer contribution-level dependencies even when given more context than the main forecasting setting provides.

\begin{table}[H]
\caption{
Task~A input and prompting variants on the held-out test set, evaluated with Gemini~3.1~Pro as the semantic matching judge. Additional citation and Related Work context improve over the claim-only setting; few-shot prompting often improves precision but does not consistently outperform richer context variants.
}
\centering
\small
\begin{tabular}{llrrr}
\toprule
Model & Setting & Recall & Precision & F1 \\
\midrule
Gemini 3.1 Pro & Main (\textbf{S1}) & 0.246 & 0.162 & 0.189 \\
Gemini 3.1 Pro & Few-shot & 0.232 & 0.226 & 0.221 \\
Gemini 3.1 Pro & +Citations (\textbf{S2}) & 0.276 & 0.192 & 0.217 \\
Gemini 3.1 Pro & +Related Work (\textbf{S3}) & 0.223 & 0.188 & 0.199 \\
\midrule
GPT-5.4 & Main (\textbf{S1}) & 0.217 & 0.123 & 0.152 \\
GPT-5.4 & Few-shot & 0.253 & 0.174 & 0.200 \\
GPT-5.4 & +Citations (\textbf{S2}) & 0.320 & 0.154 & 0.200 \\
GPT-5.4 & +Related Work (\textbf{S3}) & 0.312 & 0.171 & 0.212 \\
\midrule
Gemma-4-E4B-it & Main (\textbf{S1}) & 0.069 & 0.058 & 0.061 \\
Gemma-4-E4B-it & Few-shot & 0.079 & 0.125 & 0.094 \\
Gemma-4-E4B-it & +Citations (\textbf{S2}) & 0.114 & 0.082 & 0.092 \\
Gemma-4-E4B-it & +Related Work (\textbf{S3}) & 0.134 & 0.120 & 0.122 \\
\midrule
Gemma-4-E4B-it + LoRA & Fine-tuned & 0.107 & 0.107 & 0.101 \\
\bottomrule
\end{tabular}

\label{tab:task_a_variants_appendix}
\end{table}

\subsection{Year-Wise Results}
\label{app:year_analysis}

Table~\ref{tab:task_a_yearwise} breaks down the main Task~A setting by publication year. There is no consistent older-is-easier pattern. For the strongest models, 2025 papers are often as easy as or easier than 2023 papers. This weakens a simple memorization-based explanation of performance.

\begin{table}[H]
\caption{
Task~A main-setting test F1 by publication year under Gemini~3.1~Pro judging. Performance does not systematically improve on older papers.
}
\centering
\small
\begin{tabular}{lrrr}
\toprule
Model & 2023 F1 & 2024 F1 & 2025 F1 \\
\midrule
Gemini 3.1 Pro & 0.187 & 0.144 & 0.217 \\
Gemini Flash & 0.202 & 0.134 & 0.167 \\
GPT-5.4 & 0.149 & 0.141 & 0.164 \\
GPT-4.1 & 0.104 & 0.078 & 0.130 \\
GPT-5 Mini & 0.076 & 0.084 & 0.100 \\
Gemma-4-E4B-it & 0.063 & 0.055 & 0.066 \\
Qwen3-4B & 0.047 & 0.058 & 0.069 \\
GPT-4o & 0.067 & 0.057 & 0.045 \\
Llama-3-8B & 0.043 & 0.032 & 0.045 \\
Llama-3.1-8B & 0.045 & 0.029 & 0.020 \\
\bottomrule
\end{tabular}

\label{tab:task_a_yearwise}
\end{table}

\subsection{Decomposition by Role and Target Contribution Type}
\label{app:cont_type}

Tables~\ref{tab:task_a_role_breakdown} and~\ref{tab:task_a_claim_type_breakdown} provide a more fine-grained view of why Task~A is difficult. The strongest pattern is that models recover concrete, nameable dependencies much more reliably than abstract methodological ones. For example, model initializations and data sources often correspond to salient artifacts that are explicitly named in papers, whereas \texttt{CORE\_METHOD} contributions require reconstructing the mechanism that makes the target contribution work. This makes them harder to infer from the target claim alone.

The target-type breakdown shows a complementary pattern. Dataset, benchmark, resource, and tool targets tend to be easier because their pathways often involve visible upstream artifacts: source data, annotation protocols, pretrained models, evaluation setups, or implementation resources. Method and finding targets are harder because their enabling contributions are less likely to be recoverable as named objects and more often involve design choices, conceptual commitments, or methodological mechanisms. Together, these results suggest that current models are not simply failing to retrieve relevant scientific objects; they struggle most when pathway recovery requires explaining how a target contribution is operationally realized.

\begin{table}[H]
\centering
\small
\begin{tabular}{lrrrrrrr}
\toprule
Model & MI & DS & CF & IT & EP & TD & CM \\
\midrule
Gemini 3.1 Pro & 0.464 & 0.337 & 0.264 & 0.237 & 0.224 & 0.143 & 0.119 \\
Gemini Flash & 0.446 & 0.287 & 0.285 & 0.276 & 0.176 & 0.122 & 0.111 \\
GPT-5.4 & 0.393 & 0.257 & 0.333 & 0.158 & 0.152 & 0.204 & 0.082 \\
GPT-4.1 & 0.357 & 0.238 & 0.188 & 0.092 & 0.136 & 0.102 & 0.049 \\
Gemma-4-E4B-it & 0.089 & 0.099 & 0.090 & 0.026 & 0.064 & 0.102 & 0.045 \\
\bottomrule
\end{tabular}
\caption{
Recall by enabling-contribution role on the Task~A test set under Gemini~3.1~Pro judging. Columns abbreviate \texttt{MODEL\_INITIALIZATION} (MI), \texttt{DATA\_SOURCE} (DS), \texttt{CONCEPTUAL\_FRAMEWORK} (CF), \texttt{IMPLEMENTATION\_TOOLING} (IT), \texttt{EVALUATION\_PROTOCOL} (EP), \texttt{TRAINING\_DATA} (TD), and \texttt{CORE\_METHOD} (CM).
}
\label{tab:task_a_role_breakdown}
\end{table}

\begin{table}[H]
\centering
\small
\begin{tabular}{lrrrrrr}
\toprule
Model & Dataset & Benchmark & Tool & Resource & Method & Finding \\
\midrule
Gemini 3.1 Pro & 0.240 & 0.209 & 0.188 & 0.187 & 0.150 & 0.107 \\
Gemini Flash & 0.205 & 0.181 & 0.150 & 0.129 & 0.143 & 0.230 \\
GPT-5.4 & 0.191 & 0.160 & 0.187 & 0.171 & 0.107 & 0.152 \\
GPT-4.1 & 0.135 & 0.143 & 0.082 & 0.092 & 0.089 & 0.095 \\
Gemma-4-E4B-it & 0.109 & 0.057 & 0.036 & 0.090 & 0.036 & 0.026 \\
\bottomrule
\end{tabular}
\caption{
F1 by target contribution type on the Task~A test set under Gemini~3.1~Pro judging. Method and finding targets are harder than artifact-like targets such as datasets and benchmarks.
}
\label{tab:task_a_claim_type_breakdown}
\end{table}
\subsection{Rationale-Quality Diagnostic}
\label{app:rationale_quality}

Task~A evaluates whether models name the right enabling contributions, but a correct contribution name does not necessarily mean the model understands why that contribution is needed. We therefore run a rationale-quality diagnostic on predicted contributions that already match a gold contribution. For each matched pair, we ask whether the predicted rationale expresses the same necessity relation as the gold rationale. This diagnostic is not part of the main metric; it tests whether models recover the role of a contribution in the pathway, not only its surface identity.

Table~\ref{tab:task_a_rationale} shows that stronger generators often capture the necessity relation once they recover the right contribution. Gemini~3.1~Pro rationales match the gold rationale for 75.1\% of matched pairs, and GPT-5.4 reaches 68.8\%. In contrast, Gemma-4-E4B-it reaches 44.2\%, with nearly as many partial rationales as fully correct ones. This suggests that frontier models' Task~A successes are often substantively meaningful: when they identify the correct enabling contribution, they frequently also explain why it is necessary.

\begin{table}[H]
\caption{
Rationale-quality diagnostic on the Task~A test set. Only predicted contributions that already semantically match a gold contribution are scored. ``Same'' indicates that the predicted rationale expresses the same necessity relation as the gold rationale; ``Partial'' indicates that the rationale is related but misses an important constraint, role, or causal link.
}
\centering
\small
\begin{tabular}{lrrrrr}
\toprule
Generator & Matched pairs & Same & Partial & Different & Same rate \\
\midrule
Gemini 3.1 Pro & 173 & 130 & 39 & 4 & 0.751 \\
GPT-5.4 & 154 & 106 & 43 & 5 & 0.688 \\
Gemma-4-E4B-it & 52 & 23 & 24 & 5 & 0.442 \\
\bottomrule
\end{tabular}

\label{tab:task_a_rationale}
\end{table}
\section{Task B Additional Results}
\label{app:task_b}

This appendix reports additional Task~B results for deterministic ranking, $K=10$ evaluation, and enabling-contribution-level grounding diagnostics. These results support three conclusions from the main paper. First, gold enabling contributions consistently improve prior-work recovery across retrieval agents, rankers, and values of $K$. Second, raw model-predicted contributions remain weak search evidence, indicating that Task~A errors propagate into retrieval. Third, candidate coverage is often much higher than top-$K$ coverage, showing that both query generation and final ranking contribute to grounding failures.

Tables~\ref{tab:task_b_det_gemini} and~\ref{tab:task_b_det_gpt} report deterministic-ranking results for $K=5$ and $K=10$ on the held-out test set. The deterministic ranker uses retrieval frequency and best Semantic Scholar rank, without LLM reranking. Gold enabling contributions improve coverage substantially over claim-only retrieval for both agents. For example, with the Gemini agent at $K=5$, deterministic Coverage increases from 0.054 for claim-only retrieval to 0.237 with gold contributions; with the GPT-5.4 agent, it increases from 0.029 to 0.171. Increasing $K$ from 5 to 10 generally increases Coverage and Recall but lowers Precision, as expected when more candidate papers are returned.

Tables~\ref{tab:task_b_llm10_gemini} and~\ref{tab:task_b_llm10_gpt} report the corresponding $K=10$ LLM-reranked results. The same qualitative pattern holds: gold contributions are consistently strongest, matched predicted contributions sometimes improve over raw predicted contributions, and raw predicted contributions generally do not improve over claim-only retrieval. LLM reranking improves over deterministic ranking most clearly in the gold-contribution setting, suggesting that reranking is useful when the candidate pool already contains relevant grounding papers. However, reranking cannot recover groundings that were never retrieved, as reflected by the candidate coverage columns.

\begin{table}[H]
\caption{
Task~B deterministic-ranking results on the held-out test set with Gemini~3.1~Pro as the retrieval agent. Deterministic ranking orders candidates using retrieval frequency and best Semantic Scholar rank.
}
\centering
\small
\begin{tabular}{lrrrrr}
\toprule
Input evidence (Gemini agent) & Coverage & Recall & Precision & F1 & Cand. cov. \\
\midrule
Claim only ($K=5$) & 0.054 & 0.040 & 0.024 & 0.027 & 0.120 \\
Claim only ($K=10$) & 0.065 & 0.045 & 0.015 & 0.021 & 0.120 \\
Gold contributions ($K=5$) & 0.237 & 0.186 & 0.095 & 0.114 & 0.403 \\
Gold contributions ($K=10$) & 0.289 & 0.224 & 0.060 & 0.087 & 0.403 \\
Predicted contributions ($K=5$) & 0.034 & 0.019 & 0.016 & 0.016 & 0.122 \\
Predicted contributions ($K=10$) & 0.048 & 0.027 & 0.011 & 0.015 & 0.122 \\
Matched predicted ($K=5$) & 0.056 & 0.043 & 0.026 & 0.029 & 0.153 \\
Matched predicted ($K=10$) & 0.068 & 0.053 & 0.016 & 0.023 & 0.153 \\
\bottomrule
\end{tabular}

\label{tab:task_b_det_gemini}
\end{table}

\begin{table}[H]
\caption{
Task~B deterministic-ranking results on the held-out test set with GPT-5.4 as the retrieval agent. Deterministic ranking orders candidates using retrieval frequency and best Semantic Scholar rank.
}
\centering
\small
\begin{tabular}{lrrrrr}
\toprule
Input evidence (GPT agent) & Coverage & Recall & Precision & F1 & Cand. cov. \\
\midrule
Claim only ($K=5$) & 0.029 & 0.019 & 0.012 & 0.013 & 0.105 \\
Claim only ($K=10$) & 0.043 & 0.029 & 0.009 & 0.012 & 0.105 \\
Gold contributions ($K=5$) & 0.171 & 0.136 & 0.072 & 0.085 & 0.303 \\
Gold contributions ($K=10$) & 0.218 & 0.162 & 0.049 & 0.067 & 0.303 \\
Predicted contributions ($K=5$) & 0.009 & 0.006 & 0.008 & 0.007 & 0.067 \\
Predicted contributions ($K=10$) & 0.028 & 0.018 & 0.009 & 0.011 & 0.067 \\
Matched predicted ($K=5$) & 0.027 & 0.019 & 0.009 & 0.010 & 0.104 \\
Matched predicted ($K=10$) & 0.045 & 0.029 & 0.008 & 0.011 & 0.104 \\
\bottomrule
\end{tabular}

\label{tab:task_b_det_gpt}
\end{table}

\begin{table}[H]
\caption{
Task~B LLM-reranked results at $K=10$ on the held-out test set with Gemini~3.1~Pro as the retrieval and reranking agent.
}
\centering
\small
\begin{tabular}{lrrrrr}
\toprule
Input evidence (Gemini agent, $K=10$) & Coverage & Recall & Precision & F1 & Cand. cov. \\
\midrule
Claim only & 0.089 & 0.058 & 0.021 & 0.028 & 0.120 \\
Gold contributions & 0.362 & 0.277 & 0.077 & 0.111 & 0.403 \\
Predicted contributions & 0.068 & 0.045 & 0.016 & 0.023 & 0.122 \\
Matched predicted & 0.111 & 0.077 & 0.026 & 0.036 & 0.153 \\
\bottomrule
\end{tabular}

\label{tab:task_b_llm10_gemini}
\end{table}

\begin{table}[H]
\caption{
Task~B LLM-reranked results at $K=10$ on the held-out test set with GPT-5.4 as the retrieval and reranking agent.
}
\centering
\small
\begin{tabular}{lrrrrr}
\toprule
Input evidence (GPT agent, $K=10$) & Coverage & Recall & Precision & F1 & Cand. cov. \\
\midrule
Claim only & 0.080 & 0.054 & 0.018 & 0.025 & 0.105 \\
Gold contributions & 0.276 & 0.206 & 0.063 & 0.088 & 0.303 \\
Predicted contributions & 0.059 & 0.042 & 0.016 & 0.020 & 0.067 \\
Matched predicted & 0.085 & 0.053 & 0.016 & 0.022 & 0.104 \\
\bottomrule
\end{tabular}

\label{tab:task_b_llm10_gpt}
\end{table}

\paragraph{Contribution-level grounding diagnostic.}
Table~\ref{tab:task_b_grounding_diag} reports the enabling-contribution-level grounding diagnostic on the test set. Unlike claim-level retrieval, this setting gives the grounding agent one enabling contribution at a time and asks it either to select an acceptable prior-work grounding or mark the contribution as unmapped. This isolates grounding decisions from the full claim-level query-generation problem.

The oracle gold condition shows that grounding remains difficult even when the correct enabling contribution is provided. With gold contributions and the Gemini grounding agent, mapped accuracy is 0.268 and agent recall is 0.235. However, recall conditioned on retrieval is much higher at 0.689, indicating that the agent is often able to select the right paper once it appears in the candidate pool. This suggests that candidate retrieval is a major bottleneck. At the same time, unmapped accuracy is high at 0.824, showing that the model is relatively good at not forcing a grounding when no clean prior study exists. GPT-5.4 performs substantially worse than Gemini in the oracle condition, with mapped accuracy of only 0.057 and agent recall of 0.038.

Grounding quality drops further when the input enabling contributions come from Task~A predictions. Gemini-predicted contributions yield much lower precision and recall than gold contributions, and Gemma-4 predictions contain very few groundable contributions. This confirms that end-to-end Task~B failure reflects two compounding difficulties: predicted decompositions often fail to provide useful grounding targets, and even correct targets require effective retrieval and selection over prior work.

\begin{table}[H]
\caption{
Enabling-contribution-level grounding diagnostic on the test set. The grounder receives one enabling contribution at a time and must either select an acceptable prior-work grounding or mark it as unmapped. Recall$\,|$retrieved conditions grounding recall on at least one acceptable grounding appearing in the retrieved candidate pool.
}
\centering
\small
\resizebox{\linewidth}{!}{%
\begin{tabular}{lrrrrrr}
\toprule
Input enabling contributions & Grounder & Mapped acc. & Recall & Recall$\,|$retrieved & Precision & Unmapped acc. \\
\midrule
Gold contributions & Gemini 3.1 Pro & 0.268 & 0.235 & 0.689 & 0.476 & 0.824 \\
Gold contributions & GPT-5.4 & 0.057 & 0.038 & 0.420 & 0.134 & 0.843 \\
Predicted contributions (Gemini) & Gemini 3.1 Pro & 0.222 & 0.167 & 0.577 & 0.095 & 0.731 \\
Predicted contributions (Gemma-4) & Gemini 3.1 Pro & 0.273 & 0.159 & 0.292 & 0.046 & 0.839 \\
\bottomrule
\end{tabular}}

\label{tab:task_b_grounding_diag}
\end{table}
%%%%%%%%%%%%%%%%%%%%%%%%%%%%%%%%%%%%%%%%%%%%%%%%%%%%%%%%%%%%%%%%%%%%%%%%%%%%%%%
%%%%%%%%%%%%%%%%%%%%%%%%%%%%%%%%%%%%%%%%%%%%%%%%%%%%%%%%%%%%%%%%%%%%%%%%%%%%%%%

\end{document}